\documentclass[conference]{IEEEtran}
\IEEEoverridecommandlockouts

\usepackage[utf8]{inputenc}
\usepackage[T1]{fontenc}
\usepackage[american]{babel}
\usepackage[center]{caption}
\usepackage{graphicx}
\usepackage{listings}
\usepackage{float}
\usepackage{amsmath,amssymb,exscale}
\usepackage{blindtext, graphicx}
\usepackage{verbatim}
\usepackage{algorithm}
\usepackage{algpseudocode}
\usepackage{fancyvrb}
\usepackage{bera}
\usepackage{amsmath}
\usepackage{mathtools}
\usepackage{lipsum}
\usepackage{svg}
\usepackage{scalerel}
\usepackage{tikz}
\usetikzlibrary{svg.path}

\definecolor{orcidlogocol}{HTML}{A6CE39}
\tikzset{
  orcidlogo/.pic={
    \fill[orcidlogocol] svg{M256,128c0,70.7-57.3,128-128,128C57.3,256,0,198.7,0,128C0,57.3,57.3,0,128,0C198.7,0,256,57.3,256,128z};
    \fill[white] svg{M86.3,186.2H70.9V79.1h15.4v48.4V186.2z}
                 svg{M108.9,79.1h41.6c39.6,0,57,28.3,57,53.6c0,27.5-21.5,53.6-56.8,53.6h-41.8V79.1z M124.3,172.4h24.5c34.9,0,42.9-26.5,42.9-39.7c0-21.5-13.7-39.7-43.7-39.7h-23.7V172.4z}
                 svg{M88.7,56.8c0,5.5-4.5,10.1-10.1,10.1c-5.6,0-10.1-4.6-10.1-10.1c0-5.6,4.5-10.1,10.1-10.1C84.2,46.7,88.7,51.3,88.7,56.8z};
  }
}

\newcommand\orcid[1]{\href{https://orcid.org/#1}{\mbox{\scalerel*{
\begin{tikzpicture}[yscale=-1,transform shape]
\pic{orcidlogo};
\end{tikzpicture}
}{|}}}}

\usepackage[bookmarks=false]{hyperref}

\def\BibTeX{{\rm B\kern-.05em{\sc i\kern-.025em b}\kern-.08em
    T\kern-.1667em\lower.7ex\hbox{E}\kern-.125emX}}

\makeatletter 
\let\old@ps@headings\ps@headings 
\let\old@ps@IEEEtitlepagestyle\ps@IEEEtitlepagestyle 
\def\confheader#1{%
\def\ps@headings{%
\old@ps@headings%
\def\@oddhead{\strut\hfill#1\hfill\strut}%
\def\@evenhead{\strut\hfill#1\hfill\strut}%
}%
\def\ps@IEEEtitlepagestyle{%
\old@ps@IEEEtitlepagestyle%
\def\@oddhead{\strut\hfill#1\hfill\strut}%
\def\@evenhead{\strut\hfill#1\hfill\strut}%
}%
\ps@headings%
} 
\makeatother 

\confheader{%
} 

\begin{document}

\title{Large Language Models and Cognitive Science: A Comprehensive Review of Similarities, Differences, and Challenges}

\author{
    \IEEEauthorblockN{
        Qian Niu\textsuperscript{*,1}, 
        Junyu Liu\textsuperscript{1},
        Ziqian Bi\textsuperscript{2},
        Pohsun Feng\textsuperscript{3},
        Benji Peng\textsuperscript{4}
        Keyu Chen\textsuperscript{4},
        Ming Li\textsuperscript{4},\\
        Lawrence KQ Yan\textsuperscript{5},
        Yichao Zhang\textsuperscript{6},
        Caitlyn Heqi Yin\textsuperscript{7},
        Cheng Fei\textsuperscript{8}\\
        Tianyang Wang\textsuperscript{9}
        Yunze Wang\textsuperscript{10}
        Silin Chen\textsuperscript{11}
        Ming Liu\textsuperscript{12}\\
        Ziyuan Qin\textsuperscript{13}
        Riyang Bao\textsuperscript{13}
        Xinyuan Song\textsuperscript{13}
        Zekun Jiang\textsuperscript{14}
    }
    \IEEEauthorblockA{
        \textsuperscript{1}Kyoto University
    }
    \IEEEauthorblockA{
        \textsuperscript{2}Indiana University
    }
    \IEEEauthorblockA{
        \textsuperscript{3}National Taiwan Normal University
    }
    \IEEEauthorblockA{
        \textsuperscript{4}Georgia Institute of Technology
    }
    \IEEEauthorblockA{
        \textsuperscript{5}Hong Kong University of Science and Technology
    }
    \IEEEauthorblockA{
        \textsuperscript{6}The University of Texas at Dallas
    }
    \IEEEauthorblockA{
        \textsuperscript{7}University of Wisconsin-Madison
    }
    \IEEEauthorblockA{
        \textsuperscript{8}Cornell University
    }
    \IEEEauthorblockA{
        \textsuperscript{9}University of Liverpool, UK 
    }
    \IEEEauthorblockA{
        \textsuperscript{10}University of Edinburgh, UK
    }
    \IEEEauthorblockA{
        \textsuperscript{11}Zhejiang University
    }
    \IEEEauthorblockA{
        \textsuperscript{12}Purdue University
    }
    \IEEEauthorblockA{\textsuperscript{13}Emory University}
    \IEEEauthorblockA{\textsuperscript{14}West China Biomedical Big Data Center, West China Hospital, Sichuan University}

    \IEEEauthorblockA{
        *Corresponding Email: niu.qian.f44@kyoto-u.jp
    }
}


\maketitle

\begin{abstract}

This comprehensive review explores the intersection of Large Language Models (LLMs) and cognitive science, examining similarities and differences between LLMs and human cognitive processes. We analyze methods for evaluating LLMs cognitive abilities and discuss their potential as cognitive models. The review covers applications of LLMs in various cognitive fields, highlighting insights gained for cognitive science research. We assess cognitive biases and limitations of LLMs, along with proposed methods for improving their performance. The integration of LLMs with cognitive architectures is examined, revealing promising avenues for enhancing artificial intelligence (AI) capabilities. Key challenges and future research directions are identified, emphasizing the need for continued refinement of LLMs to better align with human cognition. This review provides a balanced perspective on the current state and future potential of LLMs in advancing our understanding of both artificial and human intelligence.

\end{abstract}


\section{Introduction}

The emergence of Large Language Models (LLMs) has sparked a revolution in artificial intelligence (AI), challenging our understanding of machine cognition and its relationship to human cognitive processes. As these models demonstrate increasingly sophisticated capabilities in language processing, reasoning, and problem-solving, they have become a focal point of interest for cognitive scientists seeking to unravel the mysteries of human cognition. This intersection of LLMs and cognitive science has given rise to a new frontier of research, offering unprecedented opportunities to explore the nature of intelligence, language, and thought.

The relationship between LLMs and cognitive science is multifaceted and bidirectional. On one hand, insights from cognitive science have informed the development and evaluation of LLMs, inspiring new architectures and training paradigms that aim to more closely mimic human cognitive processes. On the other hand, the remarkable performance of LLMs on various cognitive tasks has prompted researchers to reevaluate existing theories of cognition and consider new perspectives on how intelligence emerges from complex systems.

This review aims to provide a comprehensive overview of the current state of research at the intersection of LLMs and cognitive science. We explore the similarities and differences between LLMs and human cognitive processes, examining how these models perform on tasks traditionally used to study human cognition. We also delve into the methods developed for evaluating LLMs cognitive abilities, highlighting the challenges and opportunities in assessing AI through the lens of cognitive science. Furthermore, we investigate the potential of LLMs to serve as cognitive models, discussing their applications in various domains of cognitive science research and the insights they provide into human cognition. The review also addresses the cognitive biases and limitations of LLMs, as well as the ongoing efforts to improve their performance and align them more closely with human cognitive processes. We examine recent developments in this area, discussing the potential synergies and challenges that arise from combining these approaches.

As LLMs continue to evolve and their capabilities expand, it becomes increasingly important to critically assess their relationship with human cognition and their potential impact on cognitive science research. This review offers a balanced and comprehensive examination of these issues, presenting insights into the current state of the field. It identifies key areas for future research and discusses the challenges and opportunities at the exciting intersection of LLMs and cognitive science. By bridging AI with cognitive science, this line of inquiry promises to deepen our understanding of human cognition and inform the development of more sophisticated, ethical, and human-centric AI systems. This comprehensive and critical examination not only highlights the current achievements but also maps out a path forward in this dynamic area of study.

\section{Comparison of LLMs and Human Cognitive Processes}
LLMs have revolutionized our understanding of AI and its potential to mimic human cognitive processes. These models have shown capabilities that resemble human cognition in various tasks, including language processing, sensory judgments, and reasoning. However, despite these similarities, there are fundamental differences between LLMs and human cognitive processes that merit close examination. This section explores these similarities and differences, evaluates the methods used to assess LLMs cognitive abilities, and discusses the potential of LLMs as cognitive models. By comparing LLMs with human cognition, we can better understand the strengths and limitations of these models in emulating human thought processes.

\subsection{Similarities and differences between LLMs and human cognitive processes}

LLMs have demonstrated remarkable capabilities in various cognitive tasks, often exhibiting human-like behaviors and performance. One of the key similarities observed is in the domain of language processing. LLMs can achieve human-level word prediction performance in natural contexts, suggesting a deep connection between these models and human language processing \cite{Goldstein2020-nv}. Studies have shown that LLMs represent linguistic information similarly to humans, enabling accurate brain encoding and decoding during language processing \cite{Tuckute2024-os}. This similarity extends to the neural level, where larger neural language models exhibit representations that are increasingly similar to neural response measurements from brain imaging \cite{Mischler2024-ez}.

LLMs also demonstrate human-like cognitive effects in certain tasks. For instance, GPT-3 exhibits priming, distance, SNARC, and size congruity effects, which are well-documented phenomena in human cognition \cite{Shaki2023-zo}. Additionally, LLMs show content effects in logical reasoning tasks similar to humans, particularly in challenging tasks like syllogism validity judgments and the Wason selection task \cite{Dasgupta2022-mf}. Research has shown that LLMs can capture aspects of human sensory judgments across multiple modalities. Marjieh et al. \cite{Marjieh2023-us} demonstrated that similarity judgments from GPT models are significantly correlated with human data across six sensory modalities, including pitch, loudness, colors, consonants, taste, and timbre. This suggests that LLMs can extract significant perceptual information from language alone.

However, significant differences exist between LLMs and human cognitive processes. Humans generally outperform LLMs in reasoning tasks, especially with out-of-distribution prompts, demonstrating greater robustness and flexibility \cite{Collins2022-ce}. LLMs struggle to emulate human-like reasoning when faced with novel and constrained problems, indicating limitations in their ability to generalize beyond their training data. Lamprinidis \cite{Lamprinidis2023-js} found that LLMs' cognitive judgments are not human-like in limited-data inductive reasoning tasks, with higher errors compared to Bayesian predictors. This suggests that LLMs may not model basic statistical principles that humans use in everyday scenarios as effectively as previously thought.

Moreover, while LLMs exhibit near human-level formal linguistic competence, they show patchy performance in functional linguistic competence \cite{Mahowald2023-oz}. This suggests that LLMs may excel at surface-level language processing but struggle with deeper, context-dependent understanding and reasoning. Another notable difference lies in the memory properties of LLMs compared to human memory. Although LLMs exhibit some human-like memory characteristics, such as primacy and recency effects, their forgetting mechanisms and memory structures differ from human biological memory \cite{Janik2023-hy}. Suresh et al. \cite{Suresh2023-wb} found that human conceptual structures are robust and coherent across different tasks, languages, and cultures, while LLMs produce conceptual structures that vary significantly depending on the task used to generate responses. This highlights a fundamental difference in the stability and consistency of conceptual representations between humans and LLMs.

\subsection{Methods for evaluating LLMs cognitive abilities}

Researchers have developed various methods to evaluate the cognitive abilities of LLMs, often drawing inspiration from cognitive science and psychology. These methods aim to provide a comprehensive assessment of LLMs' capabilities and limitations in comparison to human cognition.

\begin{figure}[h]
    \centering
    \includegraphics[width=0.5\textwidth]{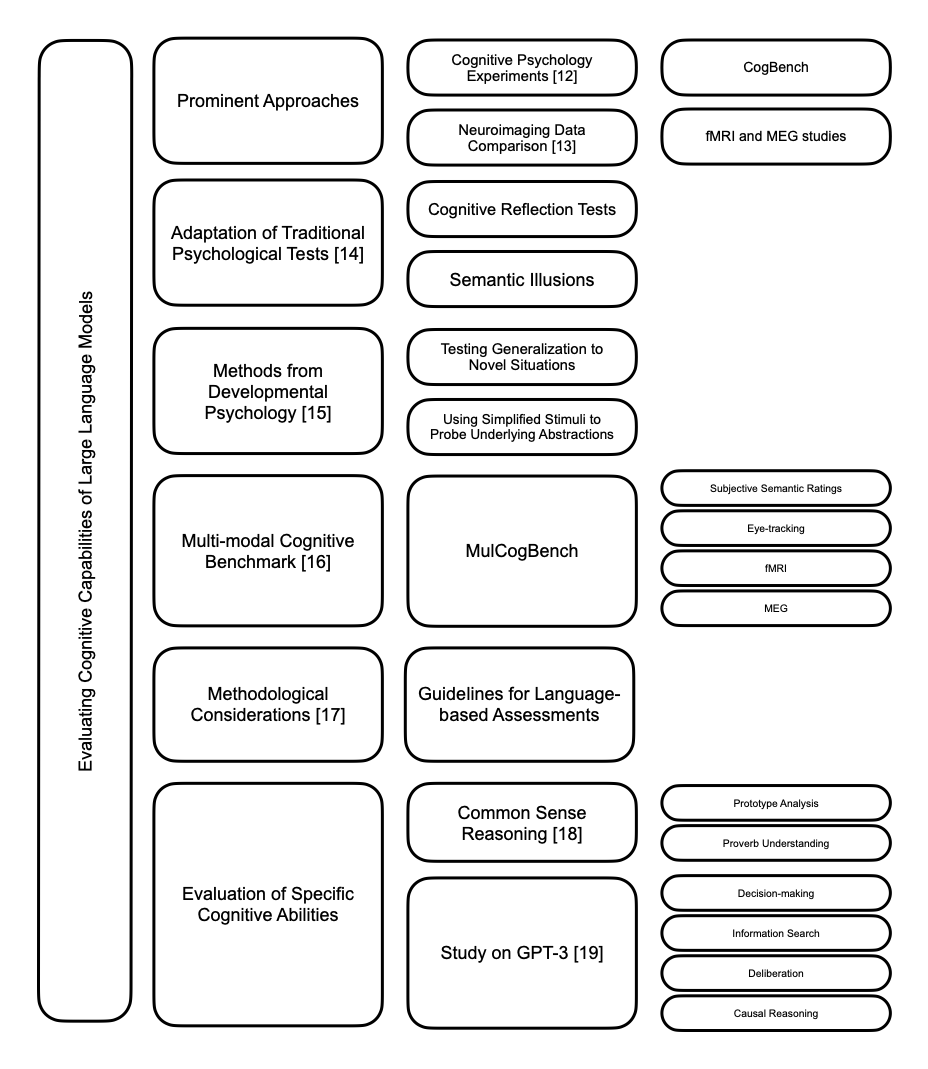}
    \caption{Evaluating Cognitive Capabilities of LLMs}
    \label{fig:InstanciaIoT}
\end{figure}

One prominent approach is the use of cognitive psychology experiments adapted for LLMs. For example, CogBench, a benchmark with ten behavioral metrics from seven cognitive psychology experiments, has been developed to evaluate LLMs \cite{Coda-Forno2024-pj}. This benchmark allows for a systematic comparison of LLMs performance across various cognitive tasks. Another method involves using neuroimaging data to compare LLMs representations with human brain activity. Studies have employed Functional magnetic resonance imaging (fMRI) and magnetoencephalography (MEG) recordings to analyze the similarity between LLMs activations and brain responses during language processing tasks \cite{Caucheteux2022-ng}. This approach provides insights into the neural-level similarities and differences between LLMs and human cognition.

Researchers have also adapted traditional psychological tests for use with LLMs. For instance, cognitive reflection tests and semantic illusions have been used to evaluate the reasoning capabilities of LLMs \cite{Hagendorff2023-ev}. These tests help reveal the extent to which LLMs exhibit human-like biases and reasoning patterns. Additionally, methods from developmental psychology have been proposed to understand the capacities and underlying abstractions of LLMs \cite{Frank2023-zr}. These approaches focus on testing generalization to novel situations and using simplified stimuli to probe underlying abstractions. 

In an effort to create more comprehensive evaluation tools, Zhang et al. \cite{Zhang2024-gr} introduced MulCogBench, a multi-modal cognitive benchmark dataset for evaluating Chinese and English computational language models. This dataset includes various types of cognitive data, such as subjective semantic ratings, eye-tracking, fMRI, and MEG, allowing for a comprehensive comparison between LLMs and human cognitive processes. Ivanova \cite{Ivanova2023-ak} provided a set of methodological considerations for evaluating the cognitive capacities of LLMs using language-based assessments. The paper highlights common pitfalls and provides guidelines for designing high-quality cognitive evaluations, contributing to best practices in AI Psychology. 

Delving deeper into specific cognitive abilities, Srinivasan et al. \cite{Srinivasan2023-vv} proposed novel methods based on cognitive science principles to test LLMs' common sense reasoning abilities through prototype analysis and proverb understanding. These methods offer new ways to assess LLMs' cognitive capabilities in more nuanced and context-dependent tasks. Binz and Schulz \cite{Binz2022-dy} used tools from cognitive psychology to study GPT-3, assessing its decision-making, information search, deliberation, and causal reasoning abilities. Their approach demonstrates the potential of cognitive psychology in studying AI and demystifying how LLMs solve tasks.

In summary, Large Language Models exhibit remarkable parallels with human cognitive processes, particularly in language and sensory tasks, yet they fall short in several critical areas, such as reasoning under novel conditions and functional linguistic competence. The diverse methodologies employed to evaluate LLMs' cognitive abilities highlight both their potential and limitations as models of human cognition. As LLMs continue to evolve, they provide a valuable tool for exploring the nature of human intelligence, but their differences from human cognitive processes must be carefully considered. Future research should aim to refine these models further, improving their alignment with human cognition and addressing the gaps that currently exist. Understanding the complex interplay between LLMs and human cognitive processes will advance both AI and cognitive science, bridging the divide between machine and human intelligence.

\section{Applications of LLMs in Cognitive Science}

The integration of LLMs into cognitive science research has opened up new avenues for understanding human cognition and developing more sophisticated AI systems. This section explores the multifaceted applications of LLMs in cognitive science, examining their role as cognitive models, their contributions to theoretical insights, and their specific applications in various cognitive domains. By synthesizing recent research, we aim to provide a comprehensive overview of the current state and future potential of LLMs in advancing our understanding of human cognition.

\subsection{LLMs as Cognitive Models}

The potential of LLMs to serve as cognitive models has gained significant attention in recent research. Studies have demonstrated that LLMs can be turned into accurate cognitive models through fine-tuning on psychological experiment data, offering precise representations of human behavior and often outperforming traditional cognitive models in decision-making tasks \cite{Binz2023-mn}. These models have shown promise in capturing individual differences in behavior and generalizing to new tasks after being fine-tuned on multiple tasks, suggesting their potential to become generalist cognitive models capable of representing a wide range of human cognitive processes. Versatility of LLMs in various cognitive domains have been explored. Wong et al. \cite{Wong2023-ro} introduced a computational framework called rational meaning construction, integrating neural language models with probabilistic models for rational inference. This approach demonstrates LLMs' ability to generate context-sensitive translations and support commonsense reasoning across various cognitive domains. Piantadosi and Hill \cite{Piantadosi2022-my} highlighted LLMs' capacity to capture essential aspects of meaning through conceptual roles, challenging skepticism about their ability to possess human-like concepts.

In the realm of language processing, Schrimpf et al. \cite{Schrimpf2020-eo} conducted a systematic integrative modeling study, revealing that transformer-based ANN models can predict neural and behavioral responses in human language processing. Their findings support the hypothesis that predictive processing shapes language comprehension mechanisms in the brain, aligning with contemporary theories in cognitive neuroscience. Kallens et al. \cite{Kallens2023-qo} demonstrated that LLMs can produce human-like grammatical language without an innate grammar, providing valuable computational models for exploring statistical learning in language acquisition and challenging traditional views on language learning. Lampinen's \cite{Lampinen2022-lo} research further challenges our understanding of human language processing, demonstrating that with minimal prompting, LLMs can outperform humans in processing recursively nested grammatical structures. This raises questions about the cognitive mechanisms underlying both human and artificial language comprehension. Nolfi \cite{Nolfi2023-ol} explored the unexpected cognitive abilities developed by LLMs through indirect processes, including dynamical semantic operations, theory of mind, affordance recognition, and logical reasoning. These findings suggest that LLMs can develop integrated cognitive skills that work synergistically, despite being primarily trained on next-word prediction tasks. This research highlights the importance of understanding these emergent capabilities in relation to human cognition. Sartori and Orrú \cite{Sartori2023-du} provided empirical evidence that LLMs perform at human levels in a wide variety of cognitive tasks, including reasoning and problem-solving. Their findings support associationism as a unifying theory of cognition and demonstrate the potential for significant impact on cognitive psychology, suggesting new avenues for modeling human cognitive processes. Li and Li \cite{Li2024-wc} proposed an intriguing duality between LLMs and Tulving's theory of memory, suggesting that consciousness may be an emergent ability based on this duality. This perspective offers a novel approach to understanding the relationship between LLMs and human cognition, potentially bridging artificial and biological intelligence research.

However, it is important to note that while LLMs can serve as plausible models of human language understanding, there are ongoing debates about the extent to which they truly capture human-like cognitive abstractions \cite{Pavlick2023-bt}. Some researchers argue that it is premature to make definitive claims about the abilities or limitations of LLMs as models of human language understanding, emphasizing the need for further empirical testing. Katzir \cite{Katzir2023-du} provided a balanced assessment of the strengths and weaknesses of LLMs, highlighting their sophisticated inductive learning capabilities while also addressing significant limitations such as opacity, data requirements, and differences from human cognitive processes. Besides, the use of LLMs as cognitive models offers new opportunities for understanding human cognition. By analyzing the internal representations and processes of these models, researchers can gain insights into potential mechanisms underlying human cognitive abilities. However, caution is necessary when interpreting these findings, as the fundamental differences in architecture and learning processes between LLMs and the human brain must be considered. Ren et al. \cite{Ren2024-hv} investigated how well LLMs align with human brain cognitive processing signals using Representational Similarity Analysis (RSA). Their findings suggest that factors such as pre-training data size, model scaling, and alignment training significantly impact the similarity between LLMs and brain activity, providing insights into how LLMs might be improved to better model human cognition.

In conclusion, while LLMs show great promise as cognitive models, further research is needed to fully understand their capabilities and limitations in representing human cognitive processes. The ongoing exploration of LLMs as cognitive models continues to provide valuable insights into both artificial and human cognition, potentially reshaping our understanding of language, reasoning, and cognitive processes.

\subsection{Insights from LLMs for cognitive science research}

LLMs have provided valuable insights for cognitive science research, challenging existing theories and offering new perspectives on human cognition. Veres \cite{Veres2022-ge} argued that while LLMs challenge rule-based theories, they do not necessarily provide deeper insights into the nature of language or cognition. This perspective highlights the need for careful interpretation of LLMs capabilities in the context of cognitive science and cautions against overinterpretation of model performance. Shanahan \cite{Shanahan2022-ad} emphasized the importance of understanding the true nature and capabilities of LLMs to avoid anthropomorphism and ensure responsible use and discourse around AI in cognitive science research. This cautionary approach underscores the need for precise language and philosophical nuance in AI discourse, particularly when drawing parallels between artificial and human cognition. Blank \cite{Blank2023-of} explored whether LLMs can be considered computational models of human language processing, discussing different interpretations and implications for future research. This work highlights the ongoing debate about whether LLMs process language like humans and the significance of this question for cognitive science, emphasizing the need for rigorous empirical investigation. Grindrod \cite{Grindrod2024-tn} argued that LLMs can serve as scientific models of E-languages (external languages), providing insights into the nature of language as a social entity. This perspective offers a novel approach to using LLMs in linguistic inquiry and cognitive science research, potentially bridging computational linguistics and sociolinguistics.

The application of LLMs in cognitive science research has opened up new avenues for exploring human behavior and decision-making processes. Horton \cite{Horton2023-bo} demonstrated the potential of using LLMs as simulated economic agents to replicate classic behavioral economics experiments. This innovative approach suggests new possibilities for using LLMs to explore human behavior and decision-making processes in cognitive science, offering a cost-effective method for piloting studies and generating hypotheses. Connell and Lynott \cite{Connell2024-ge} evaluated the cognitive plausibility of different types of language models, emphasizing the importance of learning mechanisms, corpus size, and grounding in assessing their relevance to human cognition. Their work provides a framework for critically evaluating the applicability of LLMs to cognitive modeling. Mitchell and Krakauer \cite{Mitchell2022-sf} surveyed the debate on whether LLMs understand language in a humanlike sense, advocating for an extended science of intelligence to explore diverse modes of cognition. This perspective highlights the need for a broader understanding of intelligence and cognition in the context of LLMs, encouraging interdisciplinary collaboration in AI and cognitive science research. Buttrick \cite{Buttrick2024-ky} proposed using LLMs to study cultural distinctions by analyzing the statistical regularities in their training data, offering new avenues for exploring cultural cognition and representation. This approach demonstrates the potential of LLMs as tools for investigating complex sociocultural phenomena in cognitive science. Finally, Demszky et al. \cite{Demszky2023-vo} reviewed the potential of LLMs to transform psychology by enabling large-scale analysis and generation of language data. They emphasized the need for further research and development to address ethical concerns and harness the full potential of LLMs in psychological research, highlighting both the opportunities and challenges in this emerging field.

In conclusion, LLMs have demonstrated significant potential as cognitive models and have provided valuable insights for cognitive science research. However, their limitations and the need for careful interpretation of their capabilities underscore the importance of continued research and interdisciplinary collaboration in this rapidly evolving field. Future work should focus on refining LLMs to better align with human cognitive processes, developing more rigorous evaluation methods, and addressing ethical considerations to ensure responsible and productive integration of LLMs in cognitive science research.

\subsection{Application of LLMs in specific cognitive fields}

LLMs have demonstrated significant potential in various cognitive domains, including causal reasoning, lexical semantics, and creative writing. In the realm of causal inference, Liu et al. \cite{Liu2024-do} conducted a comprehensive survey exploring the mutual benefits between LLMs and causal inference, highlighting how causal perspectives can enhance LLMs' reasoning capacities, fairness, and safety. Similarly, Kıcıman et al. \cite{Kiciman2023-zn} benchmarked the causal capabilities of LLMs, finding that they outperform existing methods in generating causal arguments across various tasks, while also noting their limitations in critical decision-making scenarios. In the field of lexical semantics, Petersen and Potts \cite{Petersen2023-jm} utilized LLMs to conduct a detailed case study of the English verb "break," demonstrating that LLM representations can capture known sense distinctions and identify new sense combinations. Their findings suggest a reconsideration of the commitment to discreteness in semantic theory, favoring a more fluid, usage-based approach. Extending to creative domains, Chakrabarty et al. \cite{Chakrabarty2023-wp} investigated the utility of LLMs in assisting professional writers through an empirical user study. Their research revealed that writers find LLMs most helpful for translation and review tasks rather than planning, while also identifying significant weaknesses in current models, such as reliance on clichés and lack of nuance.

These studies collectively underscore the diverse applications of LLMs in cognitive fields, from enhancing causal reasoning to supporting creative processes, while also highlighting areas for improvement and future research directions. In conclusion, the application of Large Language Models in cognitive science research represents a significant advancement in our ability to model and understand human cognition. LLMs have demonstrated remarkable potential as cognitive models, offering insights into language processing, reasoning, and decision-making that challenge and expand existing theories. Their versatility in addressing diverse cognitive tasks, from causal inference to creative writing, underscores their value as research tools across multiple domains of cognitive science. 

However, the integration of LLMs into cognitive research is not without challenges. Researchers must navigate issues of interpretability, ethical considerations, and the potential for overinterpretation of model capabilities. The ongoing debate about the nature of LLMs "understanding" and its relationship to human cognition highlights the need for continued critical examination and empirical investigation. As the field progresses, interdisciplinary collaboration will be crucial in refining LLMs to better align with human cognitive processes, developing more rigorous evaluation methods, and addressing ethical concerns. The future of LLMs in cognitive science research holds promise for transformative insights into the nature of intelligence, both artificial and biological, potentially bridging gaps between computational models and human cognition. By carefully leveraging the strengths of LLMs while acknowledging their limitations, researchers can continue to push the boundaries of our understanding of the mind and pave the way for more advanced AI systems that complement and enhance human cognitive abilities.

\section{Limitations and Improvement of LLMs Capabilities}

The rapid advancement of LLMs has necessitated a comprehensive evaluation of their capabilities and limitations. This section examines the cognitive biases and constraints inherent in LLMs, as well as proposed methods for enhancing their performance. By critically analyzing these aspects, researchers aim to develop more robust and reliable AI systems that can better emulate human-like cognition and language understanding.

\subsection{Cognitive biases and limitations of LLMs}

Recent studies have extensively explored the cognitive biases and limitations of LLMs. Ullman \cite{Ullman2023-pu} demonstrated that LLMs fail on trivial alterations to Theory-of-Mind tasks, suggesting a lack of robust Theory-of-Mind capabilities. Talboy and Fuller \cite{Talboy2023-xv} identified multiple cognitive biases in LLMs similar to those found in human reasoning, highlighting the need for increased awareness and mitigation strategies. Thorstad \cite{Thorstad2023-fz} advocated for cautious optimism about LLMs performance while acknowledging genuine biases, particularly framing effects. Singh et al. \cite{Singh2023-ay} investigated the confidence-competence gap in LLMs, revealing instances of overconfidence and underconfidence reminiscent of the Dunning-Kruger effect. Marcus et al. \cite{Marcus2023-fb} argued that LLMs currently lack deeper linguistic and cognitive understanding, leading to incomplete and biased representations of human language. Macmillan-Scott and Musolesi \cite{Macmillan-Scott2024-ef} evaluated seven LLMs using cognitive psychology tasks, finding that they display irrationality differently from humans and exhibit significant inconsistency in their responses. Jones and Steinhardt \cite{Jones2022-ae} presented a method inspired by human cognitive biases to systematically identify and test for qualitative errors in LLMs, uncovering predictable and high-impact errors. Smith et al. \cite{Smith2023-qr} proposed using the term "confabulation" instead of "hallucination" to more accurately describe inaccurate outputs of LLMs, emphasizing the importance of precise metaphorical language in understanding AI processes.

\subsection{Methods for improving LLMs performance}

Researchers have proposed various methods to improve LLMs performance and address their limitations. Nguyen \cite{Nguyen2023-qd} introduced the bounded pragmatic speaker model to understand and improve language models by drawing parallels with human cognition and suggesting enhancements to reinforcement learning from human feedback (RLHF). Lv et al. \cite{Lv2024-tq} developed CogGPT, an LLM-driven agent with an iterative cognitive mechanism that outperforms existing methods in facilitating role-specific cognitive dynamics under continuous information flows. Prystawski et al. \cite{Prystawski2022-ui} demonstrated that using chain-of-thought prompts informed by probabilistic models can improve LLMs' ability to understand and paraphrase metaphors. Aw and Toneva \cite{Aw2022-yo} found that training language models to summarize narratives improves their alignment with human brain activity, indicating deeper language understanding. Du et al. \cite{Du2022-fg} reviewed recent developments addressing shortcut learning and robustness challenges in LLMs, suggesting the combination of data-driven schemes with domain knowledge and the introduction of more inductive biases into model architectures.

These studies collectively highlight the importance of understanding and addressing cognitive biases and limitations in LLMs while exploring innovative methods to enhance their performance and alignment with human cognition. Future research should focus on developing more robust evaluation techniques, integrating insights from cognitive science, and creating LLMs that exhibit deeper linguistic and cognitive understanding.

In conclusion, the assessment and improvement of LLM capabilities remain critical areas of research in the field of AI. The studies reviewed in this section collectively highlight the importance of understanding and addressing cognitive biases and limitations in LLMs while exploring innovative methods to enhance their performance and alignment with human cognition. Future research should focus on developing more robust evaluation techniques, integrating insights from cognitive science, and creating LLMs that exhibit deeper linguistic and cognitive understanding. By addressing these challenges, researchers can pave the way for more advanced and reliable AI systems that can better serve human needs and contribute to various domains of knowledge and application.

\section{Integration of LLMs with Cognitive Architectures}

Recent research has explored various approaches to integrate LLMs with cognitive architectures, aiming to enhance AI systems' capabilities. This synergistic approach leverages the strengths of both LLMs and cognitive architectures while mitigating their respective weaknesses. Romero et al. \cite{Romero2023-ei} presented three integration approaches: modular, agency, and neuro-symbolic, each with its own theoretical grounding and empirical support. Kirk et al. \cite{Kirk2023-ne} explored the direct extraction of task knowledge from GPT-3 by cognitive agents, using template-based prompting and natural-language interaction. They proposed a six-step process for knowledge extraction and integration into cognitive architectures. Joshi and Ustun \cite{Joshi2024-qp} proposed a method to augment cognitive architectures like Soar and Sigma with generative LLMs, using them as prompt-able declarative memory within the architecture. González-Santamarta et al. \cite{Gonzalez-Santamarta2023-rv} integrated LLMs into the MERLIN2 cognitive architecture for autonomous robots, focusing on enhancing reasoning capabilities and human-robot interaction.

Several studies have demonstrated the potential benefits of combining LLMs with cognitive architectures in various domains. Zhu and Simmons \cite{Zhu2024-fr} presented a framework that combines LLMs with cognitive architectures to create an efficient and adaptable agent for performing kitchen tasks. Their approach demonstrated improved efficiency and fewer required tokens compared to using LLMs alone. Nakos and Forbus \cite{Nakos2024-mu} discussed the integration of BERT into the Companion cognitive architecture, showing improvements in disambiguation and fact plausibility prediction for natural language understanding tasks. Wray et al. \cite{Wray2021-jw} reviewed the capabilities of LMs for cognitive systems and proposed a research strategy for integrating LMs into cognitive agents to improve task learning and performance. They emphasized the need for effective prompting, interpretation, and verification strategies. Zhou et al.\cite{Zhou2024-du} proposed a Cognitive Personalized Search (CoPS) model that integrates LLMs with a cognitive memory mechanism inspired by human cognition to enhance user modeling and improve personalized search results.

These studies collectively demonstrate the potential of integrating LLMs with cognitive architectures to create more robust, efficient, and adaptable AI systems. However, challenges remain, including ensuring the accuracy and relevance of extracted knowledge, managing computational costs, and addressing the limitations of both LLMs and cognitive architectures. Future research directions include exploring more sophisticated integration methods, improving the efficiency of LLM-based reasoning, and investigating the application of these integrated systems in various domains.

\section{Discussion}

The intersection of LLMs and cognitive science has opened up a fascinating new frontier in AI and our understanding of human cognition. This review has highlighted the significant progress made in comparing LLMs and human cognitive processes, developing methods for evaluating LLMs cognitive abilities, and exploring the potential of LLMs as cognitive models. However, it also reveals several important areas for future research and consideration.

One of the most striking findings is the remarkable similarity between LLMs and human cognitive processes in certain domains, particularly in language processing and some aspects of reasoning. The ability of LLMs to exhibit human-like priming effects, content effects in logical reasoning, and even capture aspects of human sensory judgments across multiple modalities suggests a deep connection between these artificial systems and human cognition. This similarity extends to the neural level, with larger neural language models showing representations increasingly similar to neural response measurements from brain imaging.

However, the review also underscores significant differences between LLMs and human cognitive processes. Humans generally outperform LLMs in reasoning tasks, especially with out-of-distribution prompts, demonstrating greater robustness and flexibility. The struggle of LLMs to emulate human-like reasoning when faced with novel and constrained problems indicates limitations in their ability to generalize beyond their training data. Moreover, while LLMs exhibit near human-level formal linguistic competence, they show patchy performance in functional linguistic competence, suggesting a gap in deeper, context-dependent understanding and reasoning.

These findings highlight the need for future research to focus on enhancing the generalization capabilities of LLMs and improving their performance in functional linguistic competence. Developing methods to imbue LLMs with more robust and flexible reasoning abilities, particularly in novel and constrained problem spaces, could significantly advance their cognitive capabilities.

The review also reveals the potential of LLMs as cognitive models, with studies demonstrating that fine-tuned LLMs can offer precise representations of human behavior and often outperform traditional cognitive models in decision-making tasks. This suggests a promising avenue for using LLMs to gain insights into human cognitive processes. However, caution is necessary when interpreting these findings, as the fundamental differences in architecture and learning processes between LLMs and the human brain must be considered.

\section{Future Challenge}

Future research should focus on developing more sophisticated methods for aligning LLMs with human cognitive processes. This could involve integrating insights from cognitive science into the architecture and training of LLMs, as well as exploring novel ways to evaluate and compare LLMs performance with human cognition across a wider range of cognitive tasks.

The application of LLMs in specific cognitive fields, such as causal reasoning, lexical semantics, and creative writing, demonstrates their potential to contribute to various areas of cognitive science research. However, it also highlights the need for continued refinement and specialization of LLMs for specific cognitive domains. Future work could focus on developing domain-specific LLMs that more accurately model human cognition in particular areas of expertise.

The review also addresses the cognitive biases and limitations of LLMs, revealing that these models can exhibit biases similar to those found in human reasoning. This finding presents both challenges and opportunities. On one hand, it underscores the need for increased awareness and mitigation strategies to address these biases in AI systems. On the other hand, it offers a unique opportunity to study cognitive biases in a controlled, artificial environment, potentially leading to new insights into the nature and origins of these biases in human cognition.

The integration of LLMs with cognitive architectures represents a promising direction for future research. This approach aims to leverage the strengths of both LLMs and cognitive architectures while mitigating their respective weaknesses. Future work in this area could focus on developing more sophisticated integration methods, improving the efficiency of LLM-based reasoning within cognitive architectures, and exploring the application of these integrated systems in various real-world domains.

In conclusion, the intersection of LLMs and cognitive science offers exciting possibilities for advancing our understanding of both artificial and human intelligence. However, it also presents significant challenges that require careful consideration and further research. As we continue to explore this frontier, it is crucial to maintain a balanced perspective, acknowledging both the remarkable capabilities of LLMs and their current limitations. By doing so, we can work towards developing AI systems that not only perform well on specific tasks but also contribute to our understanding of cognition itself.

\bibliographystyle{IEEEtran}
\nocite{*}
\bibliography{citations}

\end{document}